\documentclass{article}
\pdfoutput=1
\usepackage{graphicx} 
\usepackage{subfigure}
\usepackage{natbib}
\usepackage{hyperref}
\usepackage[lined,boxed]{algorithm2e}
\usepackage[accepted]{icml2014}

\usepackage{rfmacros}
\usepackage{swmacros}

\newcommand\LRP[1]{\mathsf{LRP}_{#1}}

\newcommand\symmat[1]{\operatorname{Sym}_{#1}}  
\newcommand\sphere[1]{\mc S^{#1}}  

\begin{document}

\twocolumn[
\icmltitle{Relaxations for inference in restricted Boltzmann machines}

\icmlauthor{Sida Wang$^*$}{sidaw@cs.stanford.edu}
\icmlauthor{Roy Frostig$^*$}{rf@cs.stanford.edu}
\icmlauthor{Percy Liang}{pliang@cs.stanford.edu}
\icmlauthor{Christopher D.\ Manning}{manning@stanford.edu}
\icmladdress{Computer Science Department, Stanford University, Stanford, CA 94305, USA}

\icmlkeywords{Sampling, MAP inference, Markov random field, restricted
  Boltzmann machine, non-convex optimization}

\vskip 0.3in
]

\begin{abstract}
  We propose a randomized relax-and-round inference algorithm that
  samples near-MAP configurations of a binary pairwise Markov random
  field.  We experiment on MAP inference tasks in several restricted
  Boltzmann machines.  We also use our underlying sampler to estimate
  the log-partition function of restricted Boltzmann machines and
  compare against other sampling-based methods.
\end{abstract}

\section{Background and setup}
\label{sec:setup}

A binary pairwise Markov random field (MRF) over $n$ variables $x \in
\bits^n$ models a probability distribution $p_{\tilde A}(x) \propto
\exp(x^\T \tilde A x)$.  The non-diagonal entries of the matrix
$\tilde A \in \R^{n \by n}$ encode pairwise potentials between
variables while its diagonal entries encode unary potentials.  The
exponentiated linear term $x^\T \tilde A x$ is the \emph{negative
  energy} or simply the \emph{score} of the MRF.  A restricted
Boltzmann machine (RBM) is a particular MRF whose variables are split
into two classes, \emph{visible} and \emph{hidden}, and in which
intra-class pairwise potentials are disallowed.

\paragraph{Notation} We write $\symmat n$ for the set of symmetric $n
\by n$ real matrices, and $\sphere{k}$ to denote the unit sphere $\{ x
\in \R^k : \| x \|_2 = 1 \}$.  All vectors are columns unless stated
otherwise.

\subsection{Integer quadratic programming}

Finding the \emph{maximum a posteriori} (MAP) value of a discrete
pairwise MRF can be cast as an integer quadratic program (IQP) given
by
\begin{align}
  \maxproblemNoST {x \in \hypc^n} {x^\T A x} 
  \label{eqn:iqp}
\end{align}
Note that we have the domain constraint $x \in \hypc^n$ rather than
$\bits^n$.  We relate the two in Section~\ref{sec:hypercubes}.

\section{Relaxations}
\label{sec:problems}

Solving \eqnref{eqn:iqp} is NP-hard in general.  In fact, the MAX-CUT
problem is a special case.  Even the cases where $A$ encodes an RBM
are NP-hard in general \cite{alon2006approximating}.  We can trade off
exactness for efficiency and instead optimize a relaxed (indefinite)
quadratic program:
\begin{align}
  \maxproblemNoST { x \in [-1, 1]^n }
                  { x^\T A x }
  \label{eqn:qp}
\end{align}

Such a relaxation is \emph{tight} for positive semidefinite $A$:
global optima of the QP and the IQP have equal objective
values.\footnote{We can always ensure tightness when $A$ is not PSD,
  as in \citet{ravikumar2006quadratic}.}
Therefore \eqnref{eqn:qp} is just hard in general as \eqnref{eqn:iqp},
even though it affords optimization by gradient-based methods in place
of combinatorial search.

The following semidefinite program (SDP) is a looser relaxation of
\eqnref{eqn:iqp} obtained by extending $x$ to higher ambient
dimension:
\begin{align}
  \maxproblem { S \in \symmat{n} }
              { \tr(A S) }
              {
                & S \succeq 0            , \;\;
                  \diag(S) \leq \vec 1
              }
  \label{eqn:sdp}
\end{align}
This relaxation dates back at least to \citet{goemans1995improved},
who use it to give the first better-than-$\frac 1 2$ approximation of
MAX-CUT.

Note that, from the problem constraints, if $S$ is feasible for
\eqnref{eqn:sdp} then it must also have a factorization $S = X X^\T$
where $X \in \R^{n \by n}$ and the rows of $X$ have Euclidean norm at
most 1.  Indeed, \eqnref{eqn:sdp} is a relaxation: we can rewrite the
objective as $\tr(A X X^\T) = \tr(X^\T A X)$, then notice that
\eqnref{eqn:iqp} corresponds to a special case where the first column
of $X$ is in $\hypc^n$ and all other entries of $X$ are zero.

\subsection{Rounding}
\label{sec:rounding}

Given such $S = X X^\T$, we round it to a point $x$ feasible for the
original IQP \eqnref{eqn:iqp} by drawing a vector $g$ uniformly at
random from the unit sphere, projecting the rows of $X$ onto $g$, and
rounding entrywise.  Formally, we let $x_i = \sign(X_i^\T g)$ for $i
\in \{ 1, \ldots, n \}$.
Prior theoretical work \cite{briet2010positive,
  nesterov1998semidefinite} shows that when $A$ is positive
semidefinite this rounding is not too lossy in expectation.  Namely, we
have $\E[x^\T A x] \geq \frac 2 \pi \tr(X^\T A X)$.

\subsection{Low-rank relaxations}

The SDP relaxation \eqnref{eqn:sdp} is appealing primarily because it
is a \emph{convex} optimization problem.  Convexity, however, comes at
the cost of a loose relaxion, and with it a rounding error that may
still be too lossy in practice.  What's more, though convexity begets
computational ease in a theoretical sense, the number of variables in
the SDP is quadratic in $n$, whereas in the QP relaxation
\eqnref{eqn:qp} it is linear.  Even in simple benchmark applications
such as modeling the MNIST dataset with an RBM ($n \approx 1.3$K),
solving a semidefinite program of such size takes hours on modern
hardware.

We hence interpolate between the QP and SDP relaxations with a
sequence of optimization problems of intermediate size:
\begin{defn}
  Let $k \in \{ 1, \ldots, n \}$.
  Denote by $\LRP{k}$ the optimization problem:
  \begin{align}
    \maxproblem { X \in \R^{n \by k} }
                { \tr(X^\T A X) }
                {
                  & \| X_i \|_2 \leq 1, \; i = 1, \ldots, n
                }.
  \end{align}
  We call $k$ the \emph{width} of this optimization problem.
\end{defn}
Note that $\LRP{1}$ is equivalent to the QP \eqnref{eqn:qp}, and that
$\LRP{n}$ corresponds to the SDP \eqnref{eqn:sdp} subject to
reparameterization by $S = X X^\T$.  The $\LRP{k}$ objective is
generally non-convex; in experiments we typically seek a stationary
(locally optimal) point by projected gradient descent.  Extensive
properties of $\LRP{k}$ are studied in \citet{burer2005local}.

\subsection{Hypercube constraint reductions}
\label{sec:hypercubes}

Much of the existing literature considers RBMs over the domain $x \in
\bits^n$ instead of $x \in \hypc^n$ \cite{hinton2010practical,
  salakhutdinov2008ais}.  The two are essentially equivalent under a
linear change of variables.  Given an IQP as in \eqnref{eqn:iqp} with
objective $x^\T A x$ over $x \in \bits^n$, we can equivalently
optimize $ [\frac 1 2 (\tilde x + 1)]^\T A [\frac 1 2 (\tilde x + 1)]$
over $\tilde x \in \hypc^n$.  Conversely, in place of the objective
$\tilde x^\T A \tilde x$ over $\tilde x \in \hypc^n$, we can optimize
$(2x - 1)^\T A (2x - 1)$ for $x \in \bits^n$.

These reductions introduce cross-terms --- a linear term (of the form
$b^\T x$ for $b \in \R^n$) and a constant term (of the form $c \in
\R$).  For instance, when going from the $\bits$ domain to the $\hypc$ domain, we collect terms:
\begin{align}
  b &= \frac 1 4 (\vec 1^\T A + A \vec 1) \label{eqn:b} \\
  c &= \frac 1 4 \vec 1^\T A \vec 1
\end{align}
We may ignore $c$ as it is an additive constant that does not affect
optimization.  When optimizing over $x \in \bits^n$, $b$ can be folded
into $A$ in a new matrix
\begin{align}
  A + \diag(b).
  \label{eqn:augBits}
\end{align}
When optimizing over $x \in \hypc^n$, we can similarly fold $b$ into
$A$ by introducing a single auxiliary variable and augmenting $A$ to
\begin{align}
  \bmat{ 0 & \frac 1 2 b^\T \\ \frac 1 2 b & A }.
  \label{eqn:augHypc}
\end{align}

A caveat of these reductions is that the objective cross terms
\eqnref{eqn:b} that they introduce behave as unary coefficients
proportional to the sum of rows and columns of $A$.  Empirically, we
found that these terms, when large in magnitude, can dominate the
objective and reduce the quality of rounded solutions to the original
(unreduced) problem.

\section{Sampling}
\label{sec:sampling}

For a single relaxed solution $X$, many samples can be produced by randomized rounding.
This yields the randomized relax-and-round (rrr-MAP) algorithm, summarized in Algorithm~\ref{alg:samp}.
Given the solution $X$ whose rows are
$X_i$, we have a rounding distribution $p_X(x)$ over the corners of
the hypercube $x \in \hypc^n$ with a geometric interpretation as
follows.  Every vector $X_i$ implicitly defines a halfspace (points
$z$ such that $X_i^\T z \geq 0$).  A sign vector $x \in \hypc^n$
describes a volume of points lying within (if $x_i = 1$) or without
(if $x_i = -1$) each halfspace, and $p_X(x)$ is the proportion of the
unit sphere boundary that intersects this volume.  Formally,
\begin{align}
  p_X(x) = \Vol(\{ z \in \sphere{k} : x \circ Xz \geq 0 \}) /
  \Vol(\sphere k).
  \label{eqn:rrDist}
\end{align}
where $\circ$ denotes Hadamard (entrywise) product.

\begin{figure}[t]
  \begin{center}
    \includegraphics[width=0.5\textwidth]{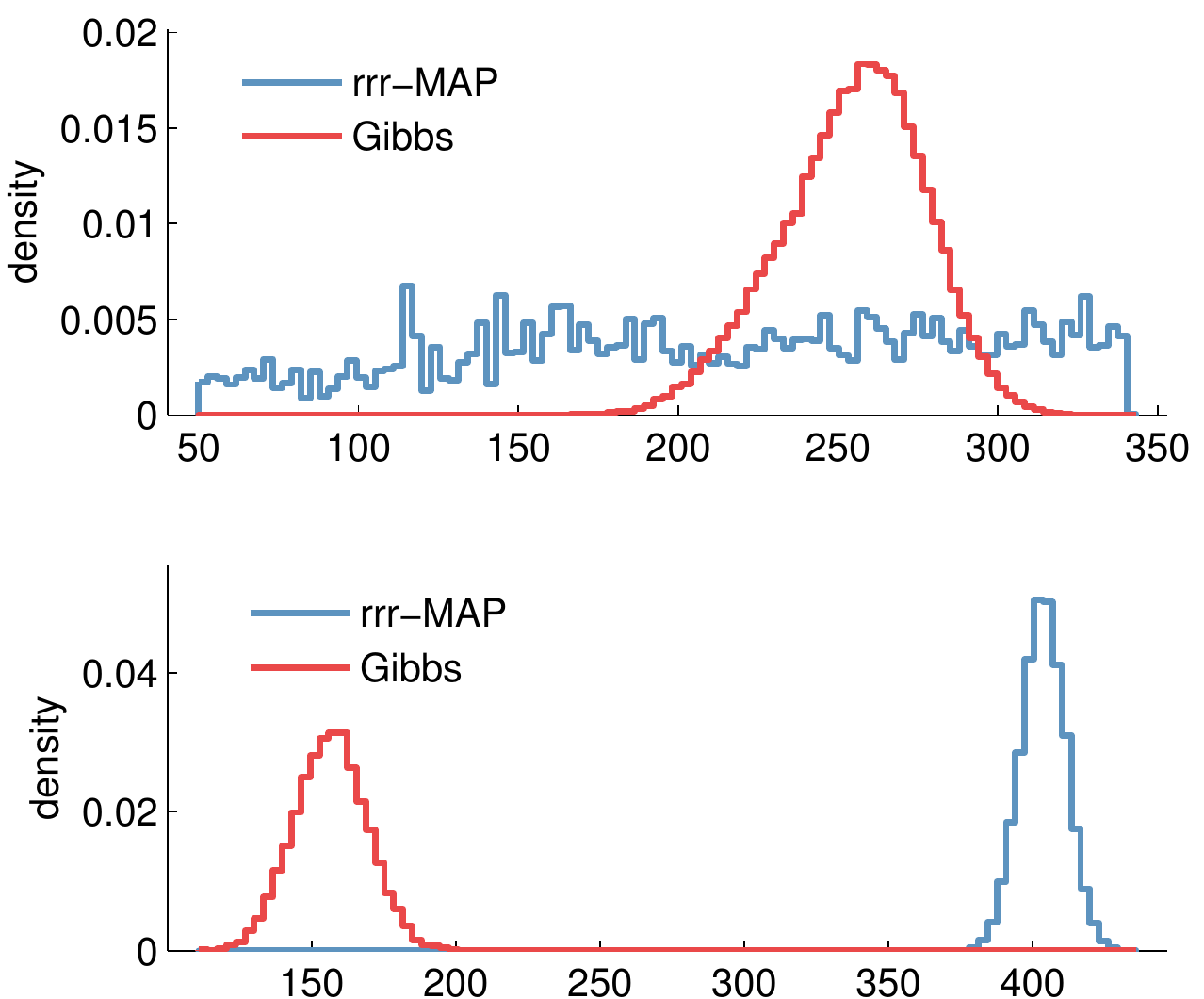}
  \end{center}
  \vspace{-1em}
  \caption{Empirical densities obtained by the rrr-MAP sampler
    (Algorithm~\ref{alg:samp}) and Gibbs sampling from two different
    RBMs.  In both cases, 10,000 samples are drawn.  Top: the
    MNIST-trained RBM of \citet{salakhutdinov2008ais}.  Bottom: a
    random RBM, with parameters sampled independently from a standard
    Gaussian.}
  \label{fig:empPdf}
\end{figure}

As shown in Figure~\ref{fig:empPdf}, this sampler produces
lower-energy samples than a mixed Gibbs sampler in both the MNIST and
random parameter settings.

\begin{algorithm}[t]
  \DontPrintSemicolon
  \SetKwInOut{Input}{Input}\SetKwInOut{Output}{Output}

  \Input{Binary pairwise MRF parameters $A$}

  \Output{Samples $\{ x^{(t)} \}_{t=1}^T$ such that $p_A(x^{(t)})$ is
    near $\max_x p_A(x)$}

  \BlankLine

  Take $X$ by optimizing $\LRP{k}$ under
  $A$\;
  \For{$t \gets 1$ \KwTo $T$}{ $g \gets$ random vector from unit
    sphere $\sphere{k}$\; $x^{(t)} \gets \sign(Xg)$\; }
  \caption{Randomized relax-and-round MAP sampler ({\bfseries rrr-MAP}).}
\label{alg:samp}
\end{algorithm}

\section{Experiments}
\label{sec:experiments}

Although our techniques are intended for general use with MRFs, we
focus entirely on RBMs in experiments.  Doing so is motivated by
special interest (largely owed to uses in feature learning) and for
sake of comparison with other sampling-based inference techniques.
The bipartite architecture of RBMs also nicely accommodates
Gibbs-based samplers.
Indeed, the RBM setting proves to be challenging for 
relaxation and rounding to do better than Gibbs variants.

An RBM with visible variables $v \in \hypc^m$ and hidden variables $h
\in \hypc^p$ fits into the above MRF framework by taking $n = m + p$
and $x = (v, h) \in \hypc^n$.
Suppose the RBM score is
\begin{align}
  v^\T W h + a^\T v + b^\T h.
  \label{eqn:rbmNegEnergy}
\end{align}
In MRF notation, we would add a single auxiliary variable and take the
augmented parameter matrix $A$ as per \eqnref{eqn:augHypc}:
\begin{align}
  A = \frac 1 2
  \bmat{
    0 & a^\T & b^\T \\
    a & 0    & W^\T \\
    b & W    & 0
  }.
\end{align}

Our experiments focus on two inference scenarios: (a) approximately
and efficiently computing the MAP, and (b) estimating the
log-partition function $\log \sum_{v,h} p_A(v,h)$.  The latter is
discussed and motivated by \citet{salakhutdinov2008ais}, and is
generally interesting as it captures the essential theoretical
hardness of RBM inference \cite{long2010restricted}.

\subsection{MAP inference}

In these benchmarks we attempt to find a low-energy configuration $x$.
We run Algorithm~\ref{alg:samp} to obtain many samples and output the
best among them.  We compare to an annealed Gibbs sampling procedure
and to an off-the-shelf IQP solver (Gurobi) that directly optimizes
\eqnref{eqn:iqp}.\footnote{\url{http://www.gurobi.com/}.}

\begin{figure}[t]
  \begin{center}
    \includegraphics[width=0.4\textwidth]{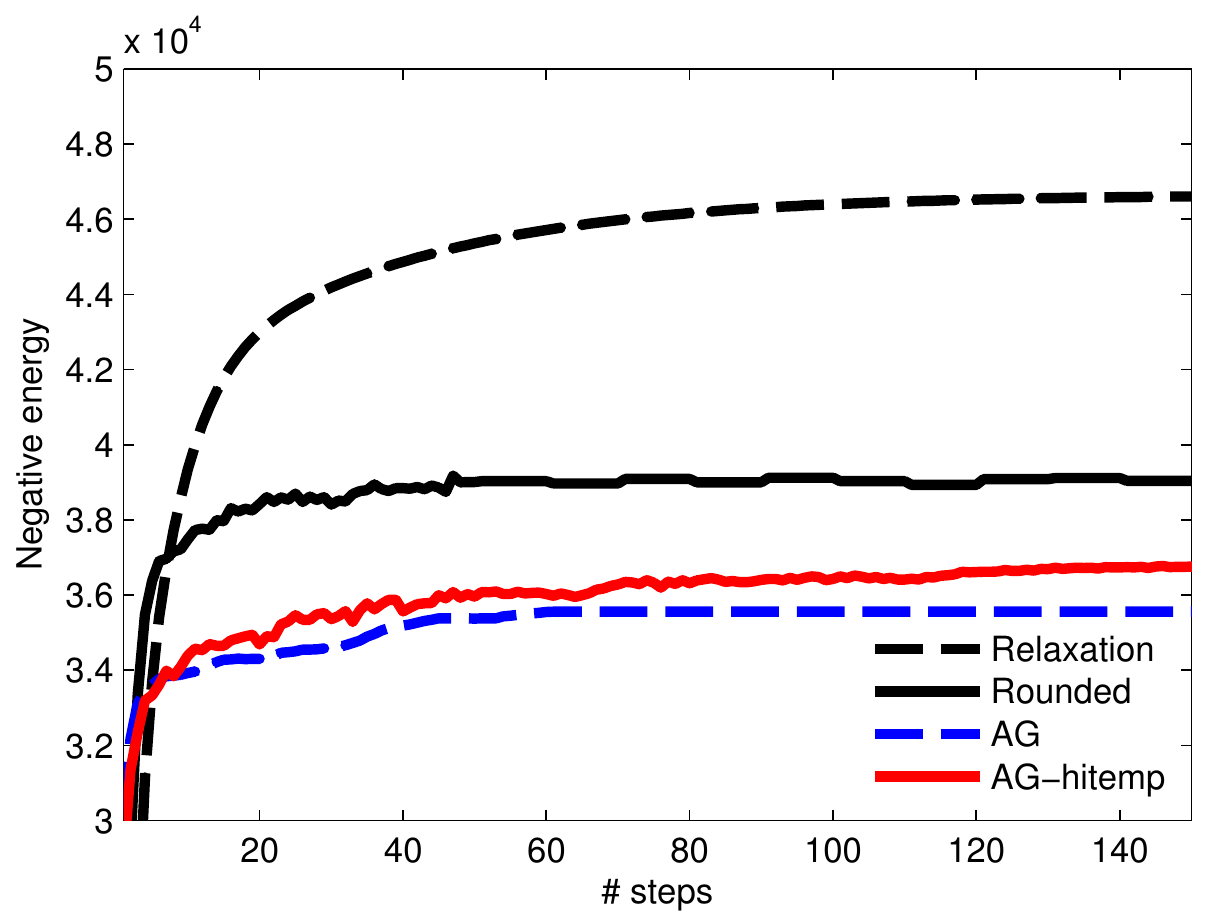}
  \end{center}
  \vspace{-1em}
  \caption{ RBM score measured across procedure steps.  The
    steps are qualitatively comparable: Gibbs requires a matrix-vector
    multiplication at every step, and $\LRP{k}$ requires a gradient
    update (dominated by matrix-vector multiplication) and projection
    onto the $L_2$ ball (\ie vector normalization). Black curves show
    the relaxed objective and the value of the best rounded sample out
    of a thousand.  Blue and red curves show annealed Gibbs, where red
    is annealing starting from a 10x higher temperature. }
  \label{fig:speed}
\end{figure}

We compare techniques across the following three RBM instances.
Results appear in Table~\ref{fig:mapTable}, and Figure~\ref{fig:speed}
illustrates convergence.
\begin{itemize}
\item \yell{MNIST.} We downloaded the weights for an RBM over the
  $\bits^n$ domain, trained by \citet{salakhutdinov2008ais} to model
  the MNIST dataset distribution.  The original model has $W \in
  \R^{784 \by 500}$ and we reduced it to an RBM over the $\hypc^n$
  domain as per Section~\ref{sec:hypercubes}.

\item \yell{Random.} We populate $W \in \R^{784 \by 500}$, $a$, and $b$ with
  independently random entries sampled from a standard Gaussian.

\item \yell{Hard.} We begin with the same type of random instance,
  then randomly select three pairs of variables, each of the form
  $(v_i, h_j)$ --- \ie one visible and one hidden.  We modify
  $W_{i,j}$ to be very large (namely, 5000) and take $a_i$ and $b_j$
  to be an order of magnitude smaller (\ie 500).  This construction is
  intended to impede a Gibbs sampler by introducing local energy
  minima.  If we initialize such a pair at $v_i = h_j = -1$, then a
  Gibbs procedure is discouraged from ever flipping the value of
  either $v_i$ or $h_j$ conditioned on the other being $-1$.
\end{itemize}

\newcommand\thd[1]{\tableCell{c}{#1}}
\newcommand\thdEnd[1]{\tableCell{c|}{#1}}
\newcommand\bfs[1]{{\bfseries #1}}
\begin{table}[t]
  \begin{center}
  \begin{tabular}{|l|rrrr|}                                     \hline
           & \thd{rrr} & \thd{AG} & \thd{rrr-AG} & \thdEnd{Gu} \\ \hline
    MNIST  & 340.29 & 377.47 & 377.39 & 319.34 \\
    Random & 22309  & 22175  & \bf{23358}  & 12939 \\
    Hard   & 40037  & 36236  & \bf{41016}  & 23347 \\
    \hline
  \end{tabular}
  \end{center}
  \vspace{-1em}
  \caption{ RBM scores found by different methods:
    \bfs{rrr} is the rrr-MAP sampler (Algorithm~\ref{alg:samp});
    \bfs{AG} is an annealed Gibbs procedure with a linear temperature schedule;
    \bfs{rrr-AG} is the annealed Gibbs procedure initialized at samples obtained from rrr-MAP;
    \bfs{Gu} is the Gurobi IQP solver.
    Executions of rrr-MAP use $\LRP{2}$ as the initial relaxation (\ie width $k=2$).
    Gurobi is given an execution time limit that is 10x that of \bfs{rrr}. }
  \label{fig:mapTable}
\end{table}

\begin{table}[t]
  \begin{center}
  \begin{tabular}{|l|rrrr|}                        \hline
             & \thd{True} & \thd{AIS} & \thd{rrr-low} & \thdEnd{rrr-IS} \\ \hline
    MNIST    & -          & 436.37    & 436.69       & 438.40 \\
    Random-S & 5127.6     & 5127.5    & 5095.7       & 5092.4 \\
    Random-L & -          & 9750.5    & 9547.7       & 9606.7 \\
    \hline
  \end{tabular}
  \end{center}
  \vspace{-1em}
  \caption{ Estimates of the RBM log-partition function $\log Z(A)$:
    \bfs{True} is the true value, when available;
    \bfs{AIS} is estimation by annealed importance sampling;
    \bfs{rrr-low} is a lower bound provided by the log-sum-exp of the energy of 10K configurations obtained by the rrr-MAP sampler;
    \bfs{rrr-IS} is estimation by importance sampling using the rrr-MAP sampler as a proposal distribution.
    Random-S indicates small $W \in \R^{784 \by 15}$. Random-L indicates $W$ of the same size as MNIST ($784 \by 500$).
    In these trials, \bfs{AIS} was run for just under twice the amount of time as \bfs{rrr-low}. }
  \label{fig:lpfTable}
  \vspace{-1em}
\end{table}

\subsection{Estimating the log-partition function}

The goal of these trials is to estimate
\begin{align}
  \log Z(A) = \log \sum_{v,h} \exp(v^\T W h + a^\T v + b^\T h).
\end{align}
The true log-partition function of large RBM instances (\eg MNIST) is
typically unknown, so we also compare results across small instances,
where the true value of $\log Z(A)$ can be computed via exhaustive
enumeration.

Throughout these benchmarks, we take advantage of the bipartite
property of RBMs to perform an analytic summation over one class of
variables.  For instance,
for a fixed $v$, we can analytically sum out $h$ in linear time:
\begin{align}
  Z(A) = \sum_v \exp(a^\T v) \prod_i \left( 1+\exp(v^\T W_i + b_i) \right).
\end{align}

We compare the following three estimation techniques.  Results are
shown in Table~\ref{fig:lpfTable}.
\begin{itemize}
\item \yell{Annealed importance sampling.}  The procedure of
  \citet{salakhutdinov2008ais}.

\item \yell{rrr-MAP sampling.}  10,000 rrr-MAP samples $\{ x^{(t)}
  \}$ are taken, and the value $\log (\sum_t \exp( {x^{(t)}}^\T A
  x^{(t)} ))$ is reported.  This is a lower bound on the true value of
  $\log Z(A)$.

\item \yell{rrr-MAP importance sampling.}  Importance sampling using
  rrr-MAP sampler as a proposal distribution.  10,000 rrr-MAP
  samples are taken and weighted by $1/p_X(x)$ (as in
  \eqnref{eqn:rrDist}) to approximate $Z$.  That is, we compute
  \begin{align}
    Z(A) \approx \E_{x \sim p_X}\left[ \frac {\exp(x^\T A x)} {p_X(x)}
    \right]
    \label{eqn:impSamp}
  \end{align}
  by estimating the right hand side with an empirical mean.  Note that
  \eqnref{eqn:impSamp} is indeed a rough approximation as $p_X$ has
  support that, for smaller $k$, is very sparse in $\hypc^n$. Using
  $k=2$, $p_X(x)$ can be computed in time $O(n)$ after a single $O(n
  \log n)$ preprocessing step.\footnote{This procedure is similar to
    the Graham scan.  The preprocessing step sorts the row vectors of
    $X$ by increasing angle.  Then $p_X(x)$ is computed for any $x$ by
    considering the row vectors in order, seeking the two consecutive
    vectors that support the cone $\{ z : x \circ Xz \geq 0 \}$.  The
    angle between these two vectors, normalized by $2\pi$, is
    $p_X(x)$.}
\end{itemize}

It is expected that sampling near the MAP (as in rrr-MAP) would help
in this estimation task whenever $\log Z(A)$ is dominated by a few
near-MAP samples.  As results show, this is perhaps a poor
assumption, and further research is needed to make rrr-MAP sampling useful
to this end.  The method does, however, remain simple to
implement, relatively efficient, and overall still comparable in
estimation quality.

\section{Conclusion}
\label{sec:conclusion}

We described an approximate MRF inference technique based on
relaxation and randomized rounding, and showed that in the RBM setting
it fares comparably to more common sampling-based methods.  When
seeking approximate MAP configurations, it succeeds in settings where
annealed Gibbs is impeded by local optima. We showed that rrr-MAP
solutions can be used to initialize local search algorithms to yield
better results than either technique finds alone.

The rrr-MAP algorithm is just as applicable more generally in MRFs,
where Gibbs sampling is less efficient than it is in the bipartite
(RBM) case.  This general setting and its surrounding theory are
examined in ongoing work.


\bibliography{main}
\bibliographystyle{icml2014}

\end{document}